\documentclass{article}
\usepackage{spconf,amsmath,graphicx,hyperref}
\usepackage{algorithm}
\usepackage{float}
\usepackage{booktabs}
\usepackage{multirow}
\usepackage{algorithmic}
\usepackage{graphicx}
\usepackage{enumitem}
\usepackage{subcaption}
\usepackage{amssymb}
\usepackage{listings}
\usepackage{booktabs}
\usepackage{multirow}

\ninept
\title{TMD-TTS: A Unified Tibetan Multi-Dialect Text-to-Speech Framework for Ü-Tsang, Amdo and Kham Speech Dataset Generation}
\name{
\begin{tabular}{c}
Yutong Liu$^{1,\dagger}$, Ziyue Zhang$^{1,\dagger}$, Ban Ma-bao$^{1}$, Renzeng Duojie$^{2}$, Yuqing Cai$^{1}$, Yongbin Yu$^{1,*}$,  \\
Xiangxiang Wang$^{1,*}$, Fan Gao$^{1}$, Cheng Huang$^{3}$, Nyima Tashi$^{1,2,*}$
\end{tabular}}
\address{$^{1}$ School of Information and Software Engineering, University of Electronic Science and \\ Technology of China, China\\
         $^{2}$ School of Information Science and Technology, Tibet University, China\\
         $^{3}$ Department of Ophthalmology, University of Texas Southwestern Medical Center, USA\\
    ybyu@uestc.edu.cn, xxwang@uestc.edu.cn, nmzx@utibet.edu.cn}
\begin{document}
%\ninept
%
\maketitle
\begin{abstract}
Tibetan is a low-resource language with limited parallel speech corpora spanning its three major dialects (Ü-Tsang, Amdo, and Kham), limiting progress in speech modeling. To address this issue, we propose TMD-TTS, a unified Tibetan multi-dialect text-to-speech (TTS) framework that synthesizes parallel dialectal speech from explicit dialect labels. Our method features a dialect fusion module and a Dialect-Specialized Dynamic Routing Network (DSDR-Net) to capture fine-grained acoustic and linguistic variations across dialects. Extensive objective and subjective evaluations demonstrate that TMD-TTS significantly outperforms baselines in dialectal expressiveness. We further validate the quality and utility of the synthesized speech through a challenging Speech-to-Speech Dialect Conversion (S2SDC) task.
\end{abstract}


\section{Introduction}
\label{sec:intro}
% Speech dataset
Tibetan, spoken by over six million people across Tibet, neighboring Chinese provinces, and parts of South Asia, is a low-resource language with three major dialects—Ü-Tsang, Amdo, and Kham—that differ substantially in phonology, lexicon, and syntax, often resulting in limited mutual intelligibility. To facilitate cross-dialect communication, speech-to-speech dialect conversion (S2SDC) is essential, yet its progress is constrained by the lack of large-scale, high-quality parallel data. Current resources remain scarce, with Zhuoma~\cite{zhuoma2022dataset} providing only about 800 parallel samples, and most datasets relying on labor-intensive manual collection. This underscores the urgent need for a TTS-based approach to synthesize Tibetan multi-dialect data.

Modern TTS systems cover a spectrum of architectures. Fully end-to-end models such as VITS~\cite{vits, vits2}, which combine cVAEs~\cite{sohn2015cvae} with normalizing flows~\cite{papamakarios2021normalizingflows}, can achieve high-fidelity synthesis but often suffer from training instability and slow inference. Diffusion-based frameworks (e.g., NaturalSpeech~\cite{naturalspeech, naturalspeech2}) and alignment-free designs (e.g., F5TTS~\cite{chen2024f5tts}) also offer strong performance but at the cost of efficiency or simplicity. In contrast, semi-end-to-end approaches such as Tacotron2~\cite{tacotron2} and FastSpeech2~\cite{fastspeech2} generate mel-spectrograms that are subsequently converted to waveforms by vocoders like HiFi-GAN~\cite{kong2020hifigan}, BigVGAN~\cite{lee2023bigvgan}, BigVSAN~\cite{shibuya2024bigvsan}, or Vocos~\cite{siuzdak2024vocos}, striking a balance between stability and controllability. Recent systems such as Matcha-TTS~\cite{mehta2024matchatts} and OT-CFM~\cite{peebles2023dit} further refine this paradigm.

Dialect modeling poses additional challenges. Research on Tibetan multi-dialect TTS remains limited: Xu et al.~\cite{xu2021} introduced a model that generates a shared mel-spectrogram and employs separate WaveNet-based vocoders for Amdo and Ü-Tsang. However, this late-stage incorporation of dialect information and the reliance on multiple heavy vocoders restrict the model’s ability to capture fine-grained dialectal distinctions.

% our study
To this end, we propose TMD-TTS, a unified Tibetan multi-dialect TTS framework that integrates dialect representations into the TTS model. Dialect embeddings are first obtained through an embedding layer and fused via a linear layer. To capture nuanced phonetic distinctions and enable precise dialectal control, we design a Dialect-Specialized Dynamic Routing Network (DSDR-Net) to replace the conventional Feedforward Network (FFN) in the Transformer architecture. DSDR-Net incorporates a conditional computation mechanism that dynamically routes information to a dialect-specific sub-network based on the input dialect ID. This allows the model to learn distinct acoustic patterns such as rhythm and intonation, offering finer-grained dialect modeling compared to shared-parameter networks.

Our main contributions are as follows:
\begin{itemize}
\item We propose TMD-TTS, the first Tibetan multi-dialect TTS framework with DSDR-Net, which significantly improves dialect consistency and captures fine-grained acoustic and linguistic variations across Tibetan dialects.
\item We construct and release TMDD, a large-scale Tibetan multi-dialect speech dataset synthesized via TMD-TTS. This provides a reproducible pipeline for high-quality dialect-rich data generation, validated through its application to a challenging S2SDC task.
\item We develop and release a comprehensive evaluation toolkit for Tibetan dialect speech synthesis, enabling standardized assessment of audio quality and dialect similarity.
\end{itemize}

\begin{figure*}[t]
  \centering
  \includegraphics[width=\textwidth]{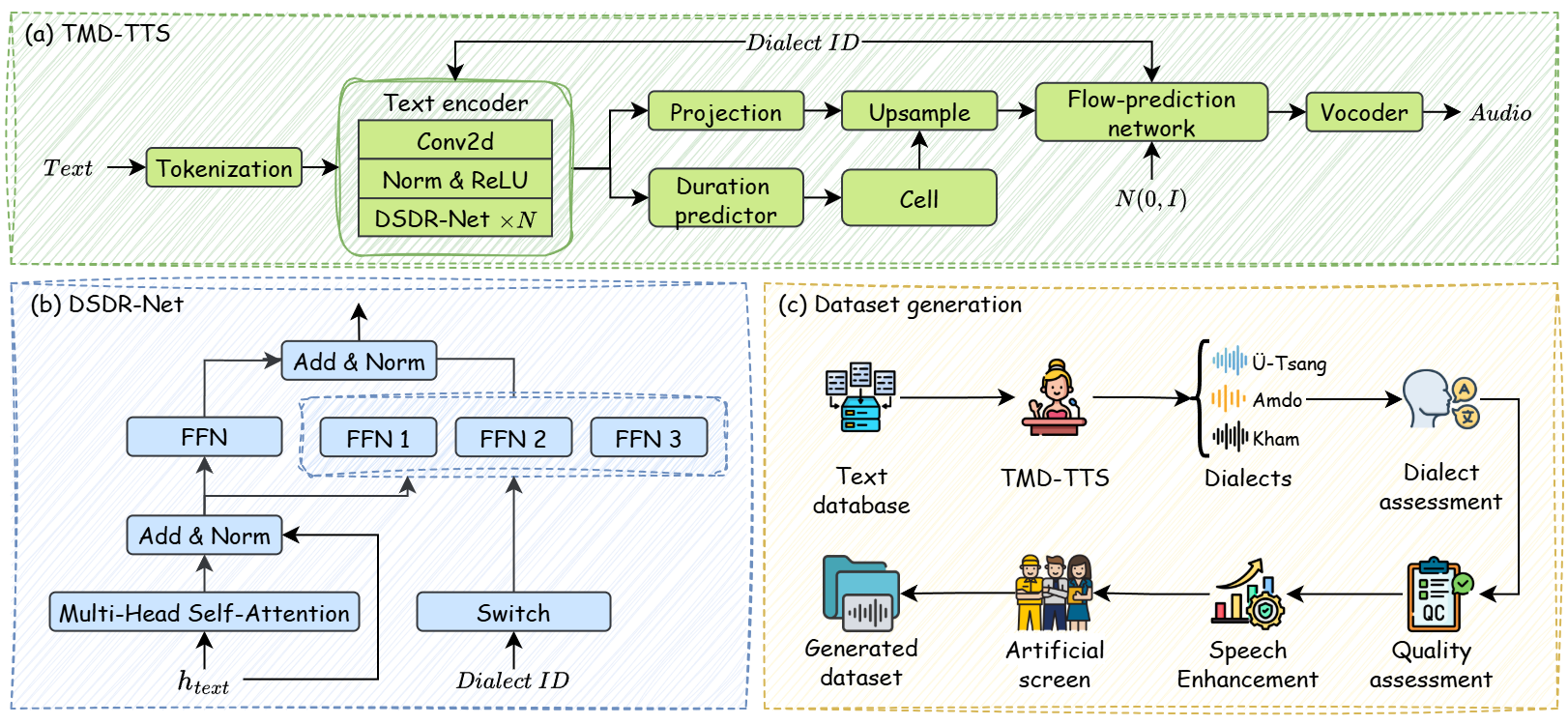}
  \caption{Overall framework of the proposed method.}
  \label{fig:TMD-TTS}
\end{figure*}

\section{Methods}
In this work, we present TMD-TTS, a unified multi-dialect TTS model built upon Matcha-TTS~\cite{mehta2024matchatts}, specifically designed for generating high-quality Tibetan dialect speech datasets. Our system addresses the limitations of prior approaches~\cite{xu2021}, such as the reliance on multi-output architectures and the need to train separate vocoders for each dialect.
The Dialect Fusion Module integrates dialect embeddings into both the text encoder and the flow prediction network, enabling dialect-aware speech synthesis. DSDR-Net introduces a conditional computation mechanism into the feed-forward network (FFN) of the Transformer architecture to provide fine-grained control over dialectal variation and accurately capture phonetic nuances across dialects.
Furthermore, we propose a complete data generation pipeline based on TMD-TTS to facilitate the large-scale synthesis of Tibetan multi-dialect speech datasets.
\subsection{TMD-TTS with DSDR-Net}
Fig.~\ref{fig:TMD-TTS}(a) illustrates the overall architecture of TMD-TTS, which synthesizes multi-dialect Tibetan speech by integrating dialect representations into the TTS pipeline. The process begins with tokenization, converting Tibetan characters into long-format tensors, followed by a text encoder. A dialect fusion module conditions the hidden representations on dialect features. Dialect processing maps each dialect $d$ to a dialect ID $did \in {0,1,2}$ and generates a normalized embedding $h_\mathrm{did}$:
\begin{equation}
\label{eq:dialect}
h_\mathrm{did} = \text{Norm}(\text{Emb}_\mathrm{did}(did))
\end{equation}
The two embeddings are concatenated, projected by a linear layer, and added to the text hidden features:
\begin{equation}
\hat{h}_\mathrm{text} = h_\mathrm{text} + \text{Linear}(h_\mathrm{did}).
\end{equation}
The duration predictor then estimates phoneme durations, guiding upsampling, while the flow-prediction network and dialect representations synthesize mel-spectrograms, which are converted into waveforms by a pretrained vocoder.

To better capture dialectal characteristics, TMD-TTS incorporates a dialect-specialized dynamic routing network (DSDR-Net, Fig.~\ref{fig:TMD-TTS}(b)) that replaces the conventional feed-forward network (FFN) in the Transformer. Given hidden features $h_{text}$, a multi-head self-attention layer produces $h_{\text{attn}}$:
\begin{equation}
h_{\text{attn}} = \text{Softmax}\left(\frac{QK^\top}{\sqrt{d_k}}\right)V,
\end{equation}
where $Q, K, V$ are projections of $h_{text}$. The information flow is then dynamically routed to a dialect-specific private FFN selected by the input dialect ID $did$:
\begin{equation}
\label{eq:ffn_private}
\text{FFN}_\mathrm{private} = \{\text{FFN}_0, \text{FFN}_1, \text{FFN}_2 \}
\end{equation}

\begin{equation}
\label{eq:switch}
\text{FFN}_\mathrm{private, did} = \text{FFN}_\mathrm{private}[did]
\end{equation}

Finally, the hidden state $h_{\text{attn}}$ is calculated by applying both the public FFN and the selected private FFN, and the outputs from the two parts are summed together:
\begin{equation}
\label{eq:dsdrnet}
\hat{h}_\text{attn} = \text{FFN}_\mathrm{public}(h_\text{attn}) + \text{FFN}_\mathrm{private, d}(h_\text{attn})
\end{equation}

\begin{table*}[t!]
  \centering
  \setlength{\tabcolsep}{3.5pt}
  \begin{tabular}{l|l|ccccccc|cc}
    \toprule
    \multirow{2}{*}{\textbf{Dialect}} & \multirow{2}{*}{\textbf{Model}} 
      & \multicolumn{7}{c|}{\textbf{Objective Metrics}} 
      & \multicolumn{2}{c}{\textbf{Subjective Metrics}} \\
    \cmidrule(lr){3-9} \cmidrule(lr){10-11}
      & & \textbf{STOI}$\uparrow$ & \textbf{PESQ}$\uparrow$ & \textbf{SI-SDR(dB)}$\uparrow$ & \textbf{DCA(\%)}$\uparrow$ & \textbf{DECS(\%)}$\uparrow$ & \textbf{DNSMOS}$\uparrow$ & \textbf{RTF}$\downarrow$ 
      & \textbf{nMOS}$\uparrow$ & \textbf{dMOC(\%)}$\uparrow$ \\
    \midrule
    \multirow{4}{*}{Ü-Tsang} 
      & SC-CNN    & 80.40 & 1.62 & 7.24 & 40.37 & 65.04 & $2.16 \pm 0.38$ & 0.036 & 2.83 & 65.14 \\
      & VITS2     & 85.72 & 2.00 & 9.88 & 39.26 & 41.91 & $2.53 \pm 0.35$ & \textbf{0.020} & 3.18 & 69.15 \\
      & Matcha-TTS & 93.84 & 2.43 & 12.32 & 65.80 & 65.20 & $2.77 \pm 0.15$ & 0.023 & 3.73 & 73.33 \\
      & TMD-TTS & \textbf{94.52} & \textbf{3.03} & \textbf{17.91} & \textbf{67.41} & \textbf{88.09} & $\textbf{2.78} \pm \textbf{0.29}$ & 0.032 & \textbf{3.83} & \textbf{76.64} \\
    \midrule
    \multirow{4}{*}{Amdo} 
      & SC-CNN & 79.90 & 1.65 & 8.25 & 59.63 & 65.04 & $2.16 \pm 0.38$ & 0.036 & 2.82 & 63.07 \\
      & VITS2 & 89.13 & 1.98 & 11.28 & 39.26 & 41.91 & $2.54 \pm 0.36$ & \textbf{0.021} & 3.20 & 70.13 \\
      & Matcha-TTS & 94.54 & 2.34 & 19.17 & 75.42 & 65.32 & $\textbf{2.79} \pm \textbf{0.13}$ & 0.023 & 3.75 & 74.56 \\
      & TMD-TTS & \textbf{94.92} & \textbf{3.13} & \textbf{21.32} & \textbf{87.78} & \textbf{79.17} & $\textbf{2.79} \pm \textbf{0.18}$ & 0.032 & \textbf{3.84} & \textbf{77.01} \\
    \midrule
    \multirow{4}{*}{Kham} 
      & SC-CNN & 76.09 & 1.47 & 5.69 & 38.52 & 19.16 & $2.01 \pm 0.30$ & 0.034 & 2.67 & 65.14 \\
      & VITS2 & 82.25 & 1.87 & 9.06 & 44.81 & 46.01 & $2.43 \pm 0.34$ & \textbf{0.021} & 3.18 & 71.11 \\
      & Matcha-TTS & 91.47 & 2.32 & 17.90 & 60.80 & 63.48 & $2.74 \pm 0.20$ & 0.022 & 3.73 & 72.16 \\
      & TMD-TTS & \textbf{93.17} & \textbf{3.05} & \textbf{21.43} & \textbf{61.11} & \textbf{67.65} & $\textbf{2.77} \pm \textbf{0.17}$ & 0.031 & \textbf{3.86} & \textbf{75.80} \\
    \bottomrule
  \end{tabular}
  \caption{\label{tab:main_result}
    Objective and subjective results on Tibetan multi-dialect TTS. 
    The highest metric is marked in \textbf{bold}. 
    ``*'' denotes the proposed model.}
\end{table*}

\subsection{Dataset Generation Pipeline}
To construct a high-quality parallel speech dataset covering the three major Tibetan dialects, we design a data generation pipeline, as illustrated in Fig.~\ref{fig:TMD-TTS}(c). In this pipeline, text is sequentially sampled from a curated database, and for each selected entry, dialectal speech is synthesized using the TMD-TTS model.

To ensure fidelity, both dialectal and perceptual quality assessments are applied. Dialect assessment is conducted by retaining only samples with a Dialect Embedding Cosine Similarity (DECS, see Section~\ref{sec:experiments}) greater than 0.8, thereby ensuring accurate dialectal representation. For perceptual quality, PESQ and DNSMOS (see Section~\ref{sec:experiments}) are employed as evaluation metrics; utterances with $PESQ < 3$ or $DNSMOS < 2.7$ are enhanced using MetricGAN+~\cite{fu2021metricgan+}. Finally, a manual screening step conducted by native speakers serves as the last stage to guarantee the reliability of the audio–text pairs, yielding the finalized dataset.

\section{Experiments}\label{sec:experiments}
\subsection{Dataset and Evaluation}
\textbf{Dataset} We constructed a 179-hour multi-dialect Tibetan speech corpus, including 44h Ü-Tsang, 45h Kham, and 90h Amdo from 1,500+ speakers. The training set contains 40k samples per dialect, with 300 samples each for validation and test.

\noindent\textbf{Setting} TMD-TTS was trained for 500k steps with Adam~\cite{kingma2017adam}, and the vocoder followed BigVGAN~\cite{lee2023bigvgan} with AdamW~\cite{loshchilov2017AdamW} and exponential decay. All experiments were conducted on two RTX 4090 GPUs. The model uses a 216-character vocabulary, 128-dim dialect embeddings, and a 192-dim DSDR-Net. For comparison, we re-implemented three baseline models, SC-CNN~\cite{sc-cnn}, VITS2~\cite{vits2}, and Matcha-TTS~\cite{mehta2024matchatts}, and extended them with the Tibetan multi-dialect TTS design from Xu et al.~\cite{xu2021}, which includes multiple BigVGAN vocoders and Wylie transliteration, to enable multi-dialect synthesis.

\noindent\textbf{Metrics} We evaluate using both subjective and objective metrics. Subjective metrics include the naturalness Mean Opinion Score (nMOS) and the dialect Mean Opinion Classification (dMOC, measured by F1 score between the predicted classifications and the labels), both evaluated by 20 native speakers. The objective evaluation includes STOI~\cite{taal2011stoi}, PESQ~\cite{rix2001pesq}, SI-SDR~\cite{leroux2019sisdr}, DNSMOS~\cite{mittal2020dnsmos}, dialect classification accuracy (DCA) obtained from a pre-trained dialect classifier inspired by metrics in Durflex-EVC~\cite{oh2025durflex}, dialect embedding cosine similarity (DECS) for measuring dialectal consistency~\cite{cai2025clap}, and the real-time factor (RTF) for inference efficiency.

\subsection{Main Result}
As shown in Table~\ref{tab:main_result}, TMD-TTS consistently outperforms all baselines across both objective and subjective metrics. For speech quality, it achieves the best performance in all three dialects, e.g., Ü-Tsang (94.52\% STOI, 3.03 PESQ, 17.91~dB SI-SDR, 2.78 DNSMOS), Amdo (94.92\% STOI, 3.13 PESQ, 21.32~dB SI-SDR, 2.79 DNSMOS), and Kham (93.17\% STOI, 3.05 PESQ, 21.43~dB SI-SDR, 2.77 DNSMOS), clearly surpassing Matcha-TTS and VITS2. In terms of dialect similarity, our model reaches up to 88.09\% DECS and 87.78\% DCA, showing clear advantages over baselines. Although its inference speed (RTF $\approx$0.031--0.032) is slightly slower than VITS2 ($\approx$0.020--0.021), it still meets the requirement for real-time synthesis. Subjective evaluations further confirm these findings, with TMD-TTS achieving the highest naturalness (nMOS 3.83/3.84/3.86) and dialect consistency (dMOC 76.64\%/77.01\%/75.80\%) across Ü-Tsang, Amdo, and Kham. Overall, these results demonstrate the superiority of TMD-TTS in both speech quality and dialectal fidelity.

\begin{table}[t]
  \centering
  \begin{tabular}{l|ccc}
    \toprule
     \textbf{Model} & \textbf{DCA(\%)} & \textbf{DECS}(\%) \\
    \midrule
    TMD-TTS & \textbf{80.25} & \textbf{78.3} \\
    \quad w/o DSDR-Net & 60.12 & 58.6 \\
    \quad w/o Dialect Fusion & 74.15 & 72.8 \\
    \quad w/o Dialect Fusion \& DSDR-Net & 33.42 & 32.2 \\
    \bottomrule
  \end{tabular}
  \caption{\label{tab:ablation_result}
    Ablation study results. The highest metric is indicated in bold.}
\end{table}

\begin{table*}[t]
  \centering
  \begin{tabular}{l|l|cccc|ccc}
    \toprule
    \textbf{Dataset} & \textbf{Dialect} & \textbf{File} & \textbf{Size} & \textbf{Dur.} & \textbf{Avg. Dur.} & \textbf{SI-SDR(dB)} & \textbf{PESQ} & \textbf{DNSMOS}\\
    \midrule
      & Ü-Tsang & 2,538 & 3.38G & 4h34m & 6.49s & 20.86 & $2.54 \pm 0.34$ & $3.55 \pm 0.06$ \\ 
    baseline & Amdo    & 1,387 & 2.56G & 2h55m & 7.57s & 23.05 & $3.11 \pm 0.20$ & $3.61 \pm 0.08$ \\
      & Kham    &   838 & 2.15G & 2h05m & 8.98s & 17.90 & $2.67 \pm 0.60$ & $3.34 \pm 0.12$ \\
      & \textbf{Total} & \textbf{4,763} & \textbf{8.09G} & \textbf{9h34m} & \textbf{7.68s} & \textbf{20.60} & $\textbf{2.77} \pm \textbf{0.38}$ & $\textbf{3.50} \pm \textbf{0.08}$ \\
    \midrule
      & Ü-Tsang & 32,714 & 6.85G & 31h55m & 3.51s & 20.92 & $3.05 \pm 0.44$ & $3.39 \pm 0.04$ \\
    TMDD & Amdo    & 32,714 &  7.79G & 34h22m & 3.78s & 18.54 & $2.84 \pm 0.43$ & $3.29 \pm 0.06$ \\
      & Kham    & 32,714 & 7.37G & 36h19m & 4.00s & 22.52 & $3.29 \pm 0.28$ & $3.36 \pm 0.06$ \\
      & \textbf{Total} & \textbf{98,142} & \textbf{22.01G} & \textbf{102h36m} & \textbf{3.76s} & \textbf{20.66} & $\textbf{3.06} \pm \textbf{0.38}$ & $\textbf{3.35} \pm \textbf{0.05}$ \\
    \bottomrule
  \end{tabular}
  \caption{\label{tab:compare_dataset} Comparison of dialectal speech statistics between the TMDD and the baseline dataset. Abbreviations: Dur (Duration), Avg (Average)}
\end{table*}

\subsection{Ablation Study}
We evaluate the contributions of dialect fusion module and the DSDR-Net in TMD-TTS (Table~\ref{tab:ablation_result}). Removing the dialect embedding (replacing it with zeros) reduces DCA from 80.25\% to 74.15\% and DECS from 78.3\% to 72.8\%, showing that explicit dialect information helps generate dialect-consistent speech. Replacing the DSDR-Net with a standard FFN drops DCA to 60.12\% and DECS to 58.6\%, highlighting its importance in modeling fine-grained dialectal variations. Removing both components further degrades performance (DCA 33.42\%, DECS 32.2\%), confirming their complementary contributions.

\begin{figure}[t]
  \centering
  \begin{subfigure}[t]{0.49\columnwidth}
    \centering
    \includegraphics[width=\linewidth]{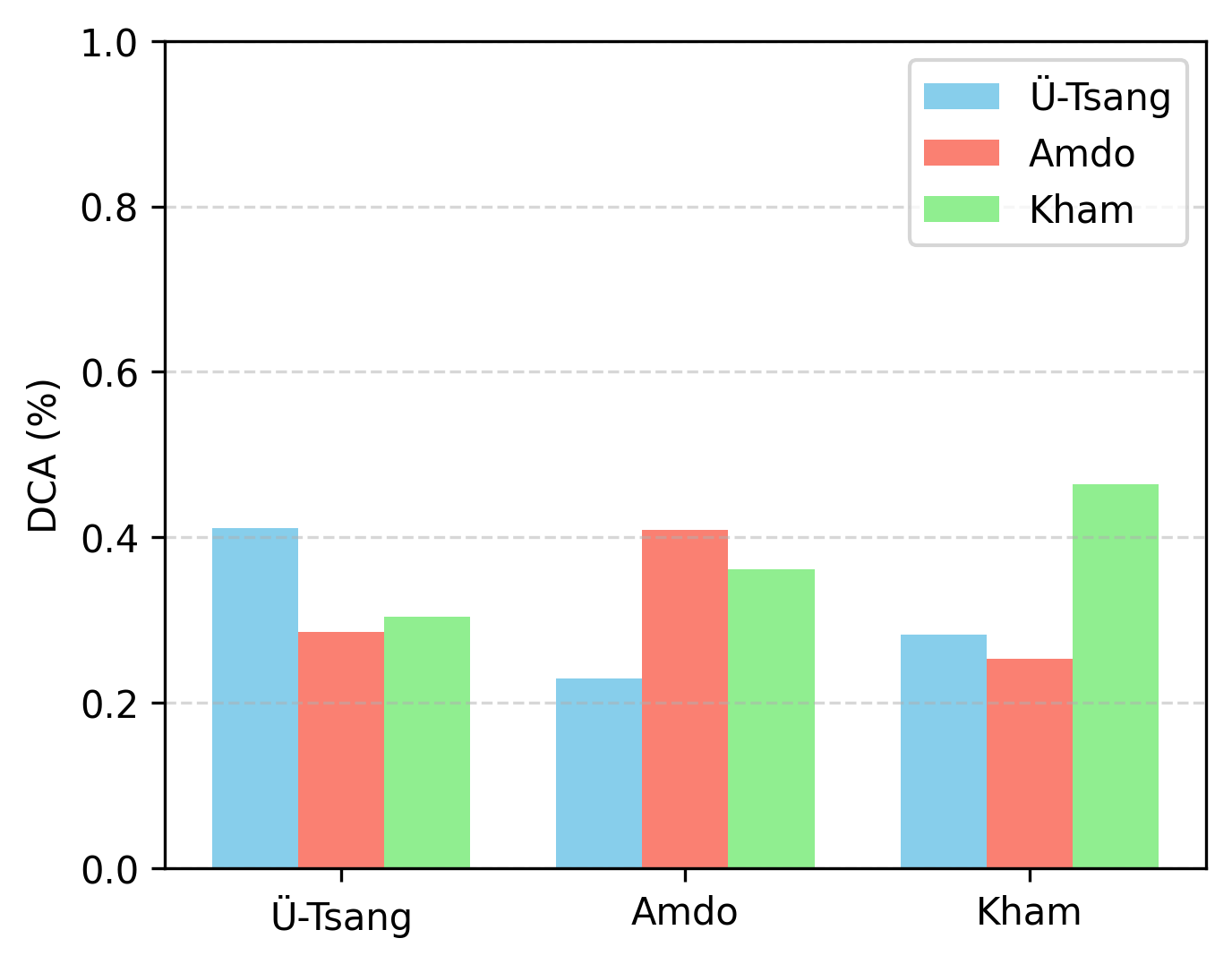}
    \caption{VITS2}
    \label{fig:dca_scores}
  \end{subfigure}
  \hfill
  \begin{subfigure}[t]{0.49\columnwidth}
    \centering
    \includegraphics[width=\linewidth]{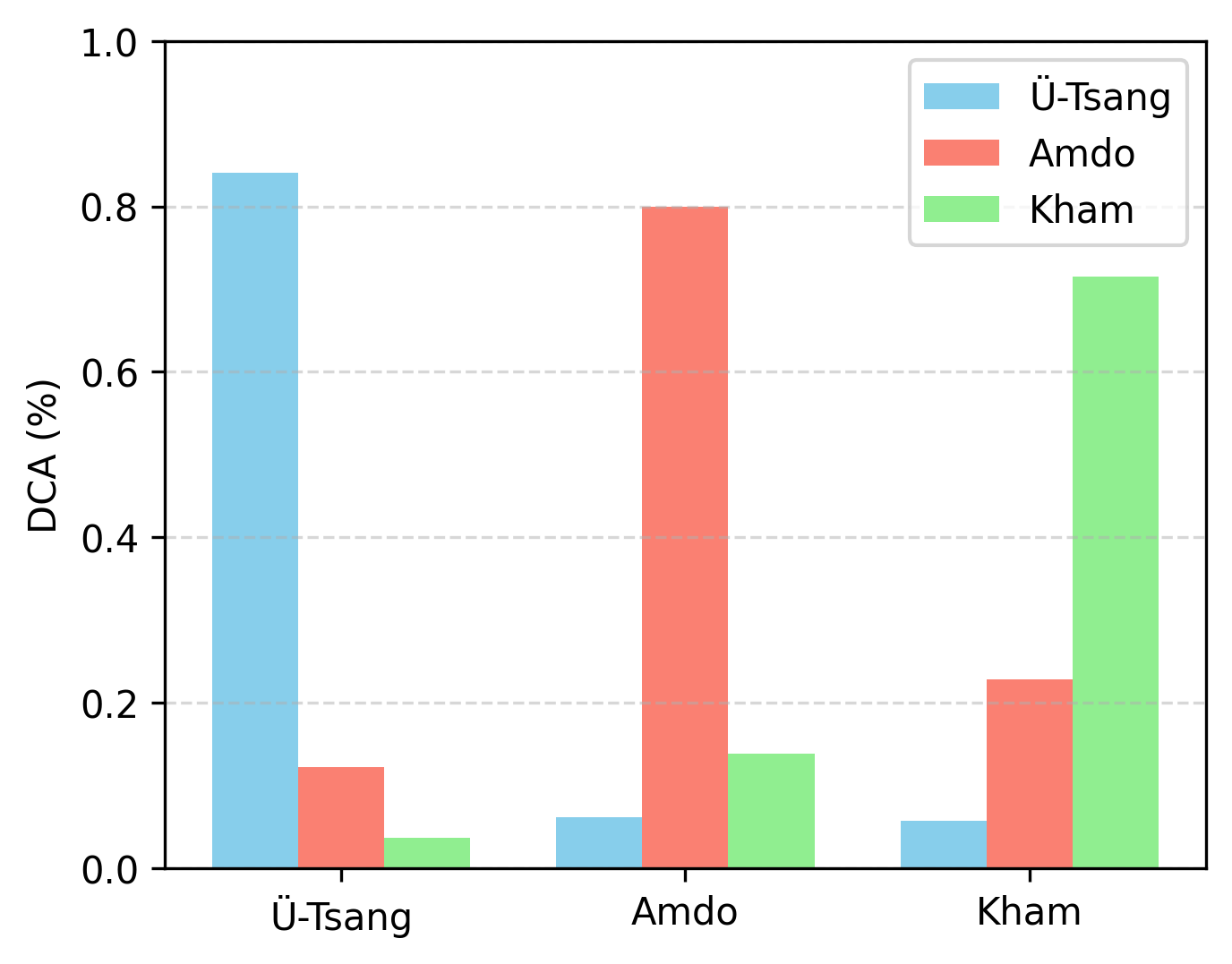}
    \caption{TMD-TTS}
    \label{fig:tsne}
  \end{subfigure}
  \caption{Visualization of the average DCA scores across three Tibetan dialects. These DCA scores reflect the classification accuracy for each component within each dialect.}
  \label{fig:dialect_analysis}
\end{figure}

\begin{figure}[t]
  \centering
  \begin{subfigure}[t]{0.49\columnwidth}
    \centering
    \includegraphics[width=\linewidth]{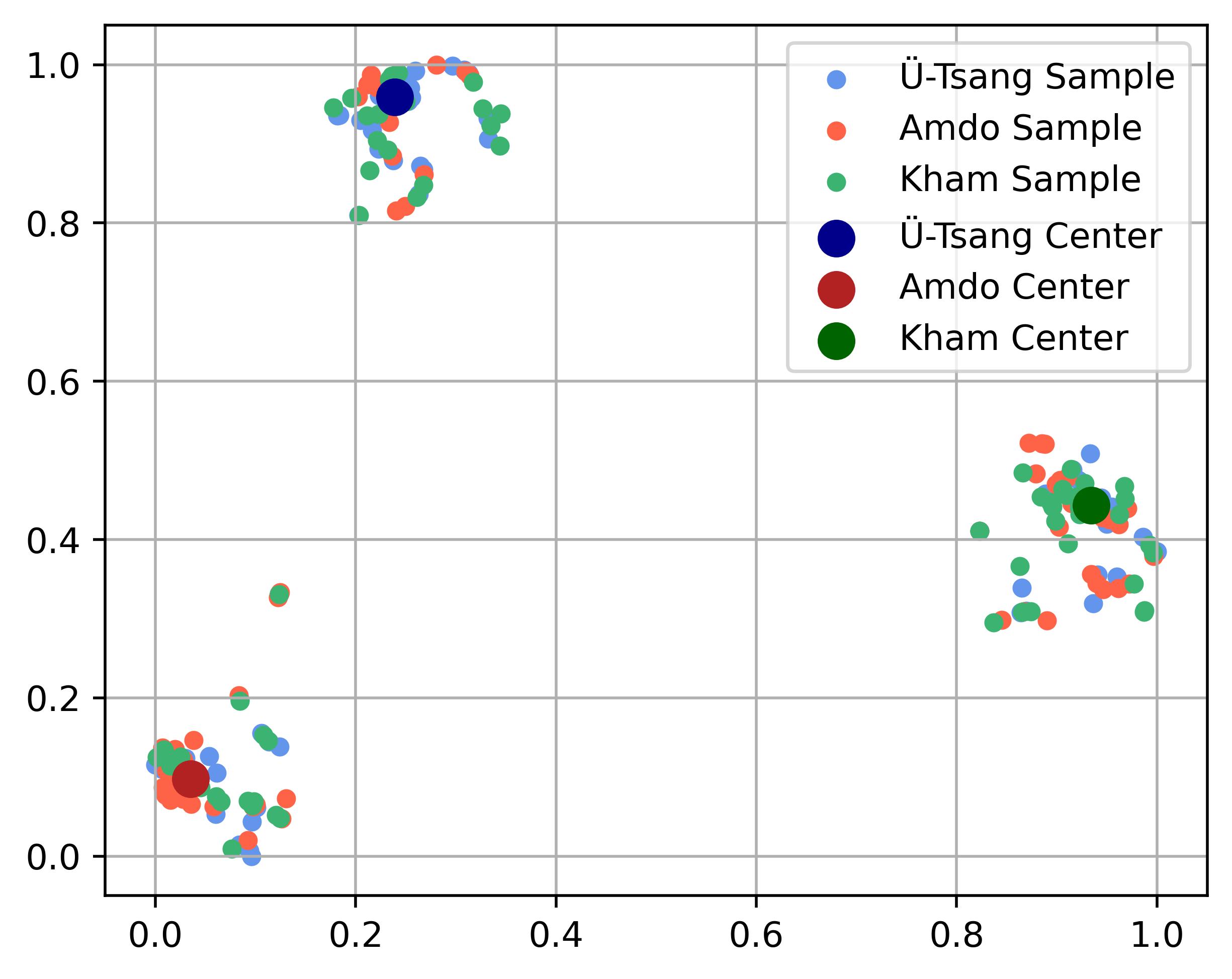}
    \caption{VITS2}
    \label{fig:t-sne_vits2}
  \end{subfigure}
  \hfill
  \begin{subfigure}[t]{0.49\columnwidth}
    \centering
    \includegraphics[width=\linewidth]{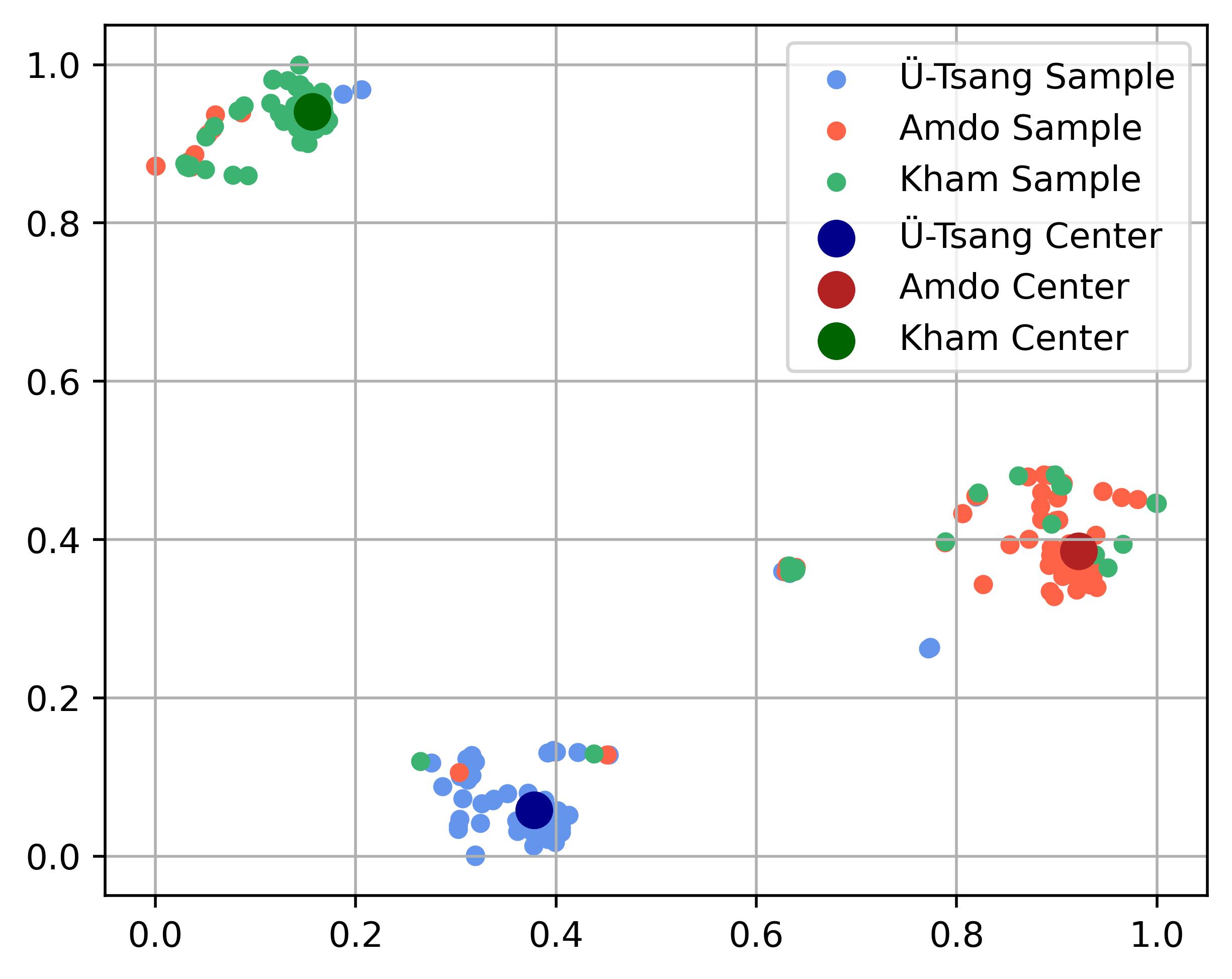}
    \caption{TMD-TTS}
    \label{fig:t-sne_TMD-TTS}
  \end{subfigure}
  \caption{Visualization of t-SNEs of dialect embedding across three Tibetan dialects.}
  \label{fig:t-sne_dialect}
\end{figure}
\subsection{Dialectal Characteristics Analysis}
To further validate the effectiveness of our proposed method in modeling dialect representations, we conducted a comparative study against the VITS2 baseline. As shown in Fig.~\ref{fig:dialect_analysis}, we use the pre-trained dialect classifier (see Section~\ref{sec:experiments}) to extract softmax-based dialectal features from each synthesized utterance, averaged over Ü-Tsang, Amdo, and Kham test samples. Compared to VITS2, our method produces speech with more salient target-dialect characteristics, demonstrating improved dialectal consistency and generalization.

Additionally, dialect embeddings were extracted using the pre-trained model for SECS computation and visualized via t-SNE (Fig.~\ref{fig:t-sne_dialect}). Each point represents a dialect embedding, with cluster centers derived from dialect embeddings. Under VITS2, embeddings are less compact and exhibit notable overlap—especially between Ü-Tsang and Amdo—indicating weaker dialect distinction. In contrast, our method achieves better cluster separation and closer alignment with the cluster centers, highlighting the effectiveness of the dialect fusion module and DSDR-Net.

\subsection{Generation and Evaluation of TMDD}
Following the dataset generation pipeline, a total of 122,700 sentences were synthesized, and 32,714 high-quality samples were selected to construct the TMDD. Compared with the baseline dataset from Zhuoma et al.~\cite{zhuoma2022dataset} (4,763 utterances, $\sim$9.5 hours), TMDD contains 98,142 utterances spanning over 102 hours, representing a 20-fold increase in sample count and an 11-fold increase in duration (Table~\ref{tab:compare_dataset}).

Objective evaluation using PESQ, SI-SDR, and DNSMOS shows that TMDD maintains high audio quality, with average PESQ $3.06 \pm 0.38$ (baseline: $2.77 \pm 0.38$) and SI-SDR $20.66$ dB (baseline: $20.60$ dB). DNSMOS results also indicate consistent perceptual quality suitable for downstream tasks.

To validate its utility, we applied TMDD to a S2SDC task using DurFlex-EVC~\cite{oh2025durflex} with BigVGAN~\cite{lee2023bigvgan} vocoding. As shown in Table~\ref{tab:s2sdc}, TMDD consistently outperforms the baseline, achieving a peak MOS of 3.63 at 22 kHz, demonstrating superior naturalness and overall quality for multi-dialect speech synthesis.

\begin{table}[t]
  \centering
  \begin{tabular}{l|c|c}
    \toprule
    \textbf{Dataset} & \textbf{Model} & \textbf{MOS}\\
    \midrule
    baseline & DurFlex-EVC + BigVGAN 16K & 3.07 \\
    baseline & DurFlex-EVC + BigVGAN 22K & 3.54 \\
    \midrule
    TMDD & DurFlex-EVC + BigVGAN 16K & 3.23 \\
    TMDD & DurFlex-EVC + BigVGAN 22K & 3.63 \\
    \bottomrule
  \end{tabular}
  \caption{\label{tab:s2sdc}
    Subjective result of DurFlex-EVC on baseline dataset and TMDD.}
\end{table}

\section{Conclusion}
In this work, we propose TMD-TTS, a unified Tibetan multi-dialect TTS model that incorporates dialect representations for multi-dialect Tibetan speech synthesis. We design a dialect fusion module and introduce DSDR-Net to better control dialectal variations. Leveraging this model, we construct and release a large-scale, parallel Tibetan multi-dialect speech dataset, TMDD, facilitating broader research in Tibetan speech synthesis and conversion. To assess the quality and utility of the generated speech, we explore the S2SDC task and conduct subjective evaluations, including naturalness and dialectal consistency. Furthermore, we develop a comprehensive evaluation toolkit specifically tailored for Tibetan speech generation tasks.

\vfill\pagebreak
\section*{Acknowledgments}
This work was supported in part by the \emph{National Science and Technology Major Project under Grant 2022ZD0116100}, in part by the \emph{National Natural Science Foundation of China under Grant 62276055}, in part by the \emph{Sichuan Science and Technology Program under Grants 23ZDYF0755, 24NSFSC5679}, in part by the \emph{Tibetan Natural Science Foundation under Grant XZ202201ZR0054G}.

\bibliographystyle{IEEEbib}
\bibliography{refs}

\end{document}